\documentclass[runningheads]{llncs}
\usepackage{graphicx}
\usepackage{comment}
\usepackage{amsmath,amssymb} 
\usepackage{color}

\usepackage{graphicx}
\usepackage{amsmath}
\usepackage{amssymb}
\usepackage{subfig}

\usepackage[breaklinks=true,bookmarks=false]{hyperref}
\usepackage[utf8]{inputenc} 
\usepackage[T1]{fontenc}    
\usepackage{hyperref}       
\usepackage{url}            
\usepackage{booktabs}       
\usepackage{amsfonts}       
\usepackage{nicefrac}       
\usepackage{microtype}      
\usepackage{graphicx}
\usepackage{algorithm}
\usepackage{algorithmic}
\usepackage{makecell}
\usepackage{mathtools}
\usepackage{multirow}
\usepackage{multicol}
\usepackage{color}
\usepackage{amssymb}
\usepackage[export]{adjustbox}
\usepackage{arydshln}
\newcommand\tab[1][0.5cm]{\hspace*{#1}}

\begin{document}
\pagestyle{headings}
\mainmatter

\title{Inducing Optimal Attribute Representations for Conditional GANs}

\titlerunning{Induce Optimal Attribs. Reprs.}
%
\author{Binod Bhattarai\inst{1} \and Tae-Kyun Kim\inst{1,2}}
\authorrunning{B. Bhattarai and T-K Kim}
%
\institute{Imperial College London, UK \and KAIST, Daejeon, South Korea \\
\email{\{b.bhattarai,tk.kim\}@imperial.ac.uk}}


\maketitle

\begin{center}
    \centering
    \includegraphics[trim={0.5cm 0cm 0.5cm 0cm},clip, width=0.15\textwidth]{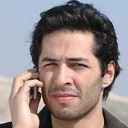}   
    \includegraphics[trim={0.5cm 0cm 0.5cm 0cm},clip,width=0.15\textwidth]{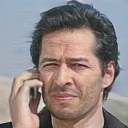}
    \includegraphics[trim={0.5cm 0cm 0.5cm 0cm},clip,width=0.15\linewidth]{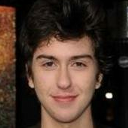}
    \includegraphics[trim={0.9cm 0cm 0.1cm 0cm},clip,width=0.15\linewidth]{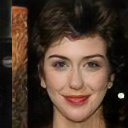}
    \includegraphics[trim={0.5cm 0cm 0.5cm 0cm},clip,width=0.15\linewidth]{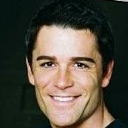}
    \includegraphics[trim={0.5cm 0cm 0.5cm 0cm},clip,width=0.15\linewidth]{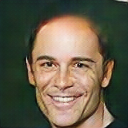}
    \captionof{figure}{Simultaneous manipulation of target facial attributes along with their auxiliary. From left to right, in each pair, original images change to Old (along with negate of black hair), Female (lipstick/makeup), and Bald (wrinkles). The attribute representations embedding relations are automatically learnt and applied to condition the generator.}
\end{center}

\begin{abstract}
 Conditional GANs (cGANs) are widely used in translating an image from one category to another. Meaningful conditions on GANs provide greater flexibility and control 
 over the nature of the target domain synthetic data. Existing conditional GANs commonly encode target domain label information as hard-coded categorical 
 vectors in the form of 0s and 1s. The major drawbacks of such representations are inability to encode the high-order semantic information of target categories 
 and their relative dependencies. We propose a novel end-to-end learning framework based on Graph Convolutional Networks to learn the attribute representations to condition 
 the generator. The GAN losses, the discriminator and attribute classification loss, 
 are fed back to the graph resulting in the synthetic images that are more natural and clearer with respect to the attributes generation. Moreover, prior-arts are mostly given priorities to condition on the generator side, not on the discriminator side of GANs. We apply 
 the conditions on the discriminator side as well via multi-task learning. We enhanced four state-of-the-art cGANs architectures: Stargan, 
 Stargan-JNT, AttGAN and STGAN. Our extensive qualitative and quantitative evaluations on challenging face attributes manipulation data set, CelebA, LFWA, and RaFD, show that the cGANs enhanced by our methods outperform by a large margin, compared to their counter-parts and other conditioning methods, in terms of both target attributes recognition rates and quality measures such as PSNR and SSIM. 

\end{abstract}
\vspace{-0.4cm}
\section{Introduction}
\begin{figure}
    \centering
    \includegraphics[trim={0cm, 0cm, 2cm, 5.5cm}, clip, width=0.3\textwidth]{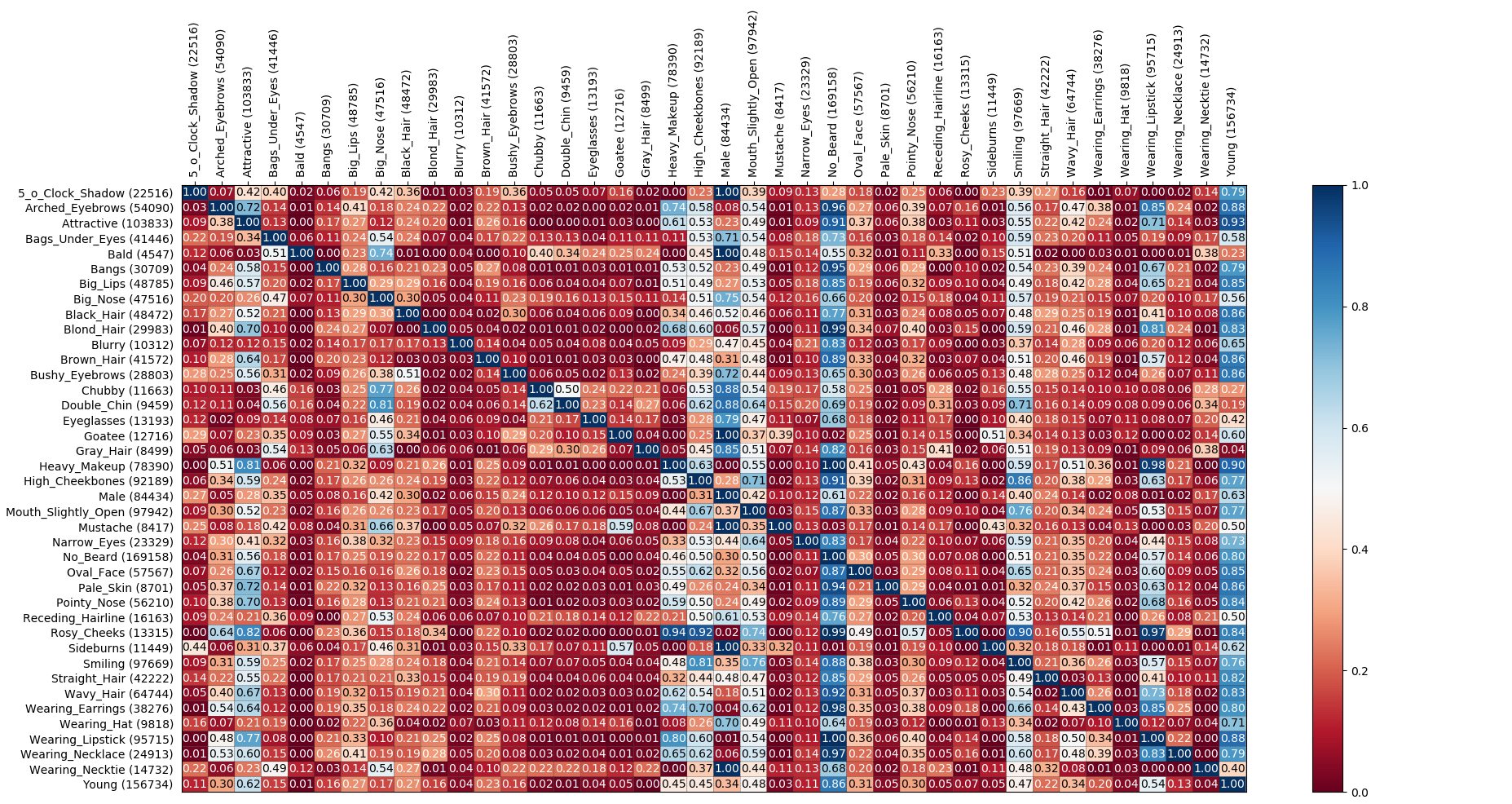}
    \caption{Co-occurrence matrix of facial attributes (zoom in the view).}
    \label{fig:coocur_matrix1}
    \vspace{-0.5cm}
\end{figure}

\vspace{-0.3cm}
\label{intro}
Someone buying bread is likely to buy butter, blue sky comes with a sunny day. Similarly, some of the attributes
of the faces co-occur more frequently than others. Fig.~\ref{fig:coocur_matrix1} shows co-occurring probabilities 
of facial attributes. We see some set of attributes such as \textit{wearing lipsticks}
and \textit{male} are least co-occurring (0.01) and \textit{male} and \textit{bald} are highly co-related (1.0). 

Face attribute manipulation using GAN~\cite{stargan_cvpr2018,he2019attgan,liu2019stgan,shen2017learning,zhang2018generative,chen2018facelet,xiao2018elegant,cao2019biphasic,karras2019style} is
one of the challenging and popular research problem. Since the advent of  conditional GAN~\cite{mirza2014conditional},
several variants of conditional GANs (cGANs) have been proposed. For conditioning the GAN, existing methods rely on target domain one-hot vectors~\cite{stargan_cvpr2018,he2019attgan,lample2017fader,miyato2018cgans}, 
synthetic model parameters of target
attributes~\cite{gecer2018eccv}, facial action units~\cite{pumarola2018ganimation}, or key point
landmarks~\cite{mirza2014conditional}, to mention a few of them. 
Recently, ~\cite{liu2019stgan} proposed to use the difference of one-hot vectors corresponding to the target 
and source attributes. This trick \textit{alone} boosts 
Target Attributes Recognition Rate (TARR) on synthetic data compared to~\cite{stargan_cvpr2018} by a large margin. 
Another recent study on GAN's~\cite{odena2018generator} identified conditioning on GAN is co-related 
with its performance. The major limitation of existing cGANs for arbitrary multiple face attributes 
manipulation is~\cite{stargan_cvpr2018,he2019attgan,liu2019stgan,perarnau2016icgan,lample2017fader} are: hard coded 1 and 0 form, 
treating every attribute equally different and ignoring the co-existence of the attributes.
In reality, as we can see in Figure~\ref{fig:coocur_matrix1}, some attributes are more co-related than others.
Moreover, the existing methods are giving less attention to conditioning on the discriminator side except
minimising the cross-entropy loss of target attributes.

Another recent work~\cite{liu2019attribute} identified the problem of artefacts on synthetic examples due to unnatural transition
from source to target. This problem arises due to the ignorance of existing GANs regarding the co-existence of certain sub set of 
attributes. To overcome this, they propose a hard-coded method to condition both target attribute (aging) and its associated 
attributes (gender, race) on generator and also on discriminator in order to faithfully retain them after translation.
However, this approach is limited to a single attribute and infeasible to hard code such rules in the case like ours 
where multiple arbitrary attributes are manipulated simultaneously. Recent study on GAN~\cite{chen2019self} identifies the forgetting 
problem of discriminator due to the non-stationary nature of the data from the generator. Applying a simple structural 
identification loss (rotation angle) helps to improve the performance and stability of GAN. 

To address the above mentioned challenges of cGANs, we investigate few continuous representations including
semantic label embedding (word2vec)~\cite{mikolov2013distributed}, attributes model parameters (attrbs-weights)
(see Sec.~\ref{Sec: compared methods}). 
The advantages of conditioning with such representations instead of  0s and 1s form 
are mainly two folds: i) carries high-order semantic information, ii) establishes relative relationship between the attributes.
These representations are, however, still sub-optimal and less natural as these are computed offline and also do not capture 
simultaneous existing tendency of different face attributes. 
Thus, we propose a novel conditioning method for cGAN to induce higher order semantic representations of target attributes, 
automatically embedding inter-attributes co-occurrence and also sharing the information based on degree of interaction. Towards this goal, 
we propose to exploit the attributes co-occurrence probability and apply Graph Convolutional Network (GCN)~\cite{kipf2016semi}
to condition GAN. GCN is capable of generating dense vectors,
distilling the higher dimensional data and also capable of convolving the un-ordered data~\cite{hamilton2017representation}.
The conditioning parameters i.e. GCN are optimised via the discriminator and attribute classification losses of GAN in an end-to-end fashion. In order to maintain such semantic structural relationship of the attributes at the discriminator side as well, we adapted on-line multitask learning objectives~\cite{cavallanti2010linear} constrained by the same co-occurrence matrix. The experiments show that the proposed method substantially improve state-of-the-art cGAN methods and other conditioning methods in terms of target attributes classification rates and PSNR/SSIM. The synthesised images by our method exhibit associated multi-attributes and clearer target attributes. Details of the method are explained in Sec.~\ref{proposed_method}, following the literature review in Sec.~\ref{related_works}, and experimental results and conclusions are shown in Sec.~\ref{experiments} and Sec.~\ref{conclusions}.
\section{Related Works}
\label{related_works}
\noindent \textbf{Conditional GANs.} After the seminal work from Mirza et al.~\cite{mirza2014conditional} on Conditional 
GANs (cGANs), several variants such as, conditioning target category labels in addition to the input~\cite{odena2018generator,chen2016infogan,stargan_cvpr2018,he2019attgan,liu2019stgan,zhang2017age}, semantic text or 
layout representations conditioning~\cite{reed2016generative,zhang2017stackgan,hong2018inferring,zhao2019image}, image
conditioning~\cite{isola2017image,huang2018multimodal,liu2017unsupervised}, facial landmarks~\cite{zakharov2019few},
have been proposed to solve different vision tasks. These works highlight the importance of semantic representations of target domain as 
a condition. 
In this work, we focus on conditioning target category labels especially by continuous and semantic representations.
~\cite{kaneko2017generative} proposes multiple strategies for random continuous representation to encode target label but limits
to a single attribute.
~\cite{ding2018exprgan} extended similar approaches for arbitrary attributes.
~\cite{kaneko2018generative} proposes to use decision tree to generate hierarchical codes to control the target attributes.
These are some of the works related to ours in terms of inducing continuous representations.
Recent works on cGANs for face attribute manipulations~\cite{stargan_cvpr2018,perarnau2016icgan,lample2017fader,he2019attgan} 
encodes in the form of 0s and 1s or their difference~\cite{liu2019stgan}. These representations are hard-coded. 
STGAN~\cite{liu2019stgan} also proposes conditional control of the information flow from source to target in the intermediate layers of GANs. This is one of the closest works in terms of the adaptive conditioning target information. 
Other cGANs, such as StyleGAN~\cite{karras2019style} propose to condition on the intermediate layers of 
the generator. Progressive GAN~\cite{karras2017progressive} proposed to gradually increase the parameters
of the generator and discriminator successively to generate high quality images. Our method is orthogonal to this line of methods, and can be extended to these works for a higher quality.
Recently, attribute aware age progression GAN~\cite{liu2019attribute} proposes to condition both associated attributes and target 
attributes at both generator and discriminator side. This work is closest in terms of conditioning at both the sides and retaining 
the auxiliary attributes in addition to target attribute. 
This approach limits to single attribute manipulation i'e aging, whereas, our method supports multiple attributes.
Also their method is hard-coded whereas our method is automatic and directly inferred from the co-occurrence matrix. \\
\noindent \textbf{Graph Convolutional Network (GCN).} Frameworks similar to~\cite{kipf2016semi} are popular for several tasks
including link prediction~\cite{grover2016node2vec}, clustering~\cite{perozzi2014deepwalk}, node classification~\cite{tang2015line}.
Recent works on image classification~\cite{chen2019multi} and  face attributes classification~\cite{nian2019facial}
propose to use GCN to induce more discriminative representations of attributes by sharing information between 
the co-occurring attributes. 
Unlike these works, we propose to apply GCN to induce such higher-order representations of target categories for the generative neural networks and optimise it via end-to-end adversarial learning. 
To the best of our knowledge, this is the first work to use such embedding as conditions in cGANS.
For more details on the work based on Graph Convolutional Networks, we suggest reader to refer to~\cite{zhou2018graph}. \\
\noindent \textbf{Regularizing/Conditioning the Discriminator.}
Conditioning on the discriminator side has been shown useful in generating more realistic and diverse
images~\cite{miyato2018cgans,miyato2018spectral,chen2019self,odena2017conditional}.~\cite{odena2018generator} maximises the distribution 
of target label in addition to source distribution to improve the quality of synthetic images.
\cite{chen2019self} introduced rotation loss on the discriminator side to mitigate the forgetting problem of 
the discriminator. Projecting the target conditional vector to the penultimate representation of the input at 
discriminator side~\cite{miyato2018cgans} substantially improved the quality of synthetic images. Another work on Spectral 
normalisation of weight parameters~\cite{miyato2018spectral} of every layer of the discriminator stabilises the training 
of GANs. Recent works on face attribute manipulations~\cite{stargan_cvpr2018,he2019attgan,liu2019stgan} minimise 
the target label cross entropy loss on discriminator. 
In this work, we introduce conditioning of the discriminator with multi-task learning framework 
while minimising the target attributes cross entropy loss. 

\section{Proposed Method}
\label{proposed_method}
\subsection{Overview on the Pipeline}

\begin{figure*}
    \centering
    \includegraphics[trim= 0.2cm 8cm 0cm 0.0cm, clip, width=0.93\textwidth]{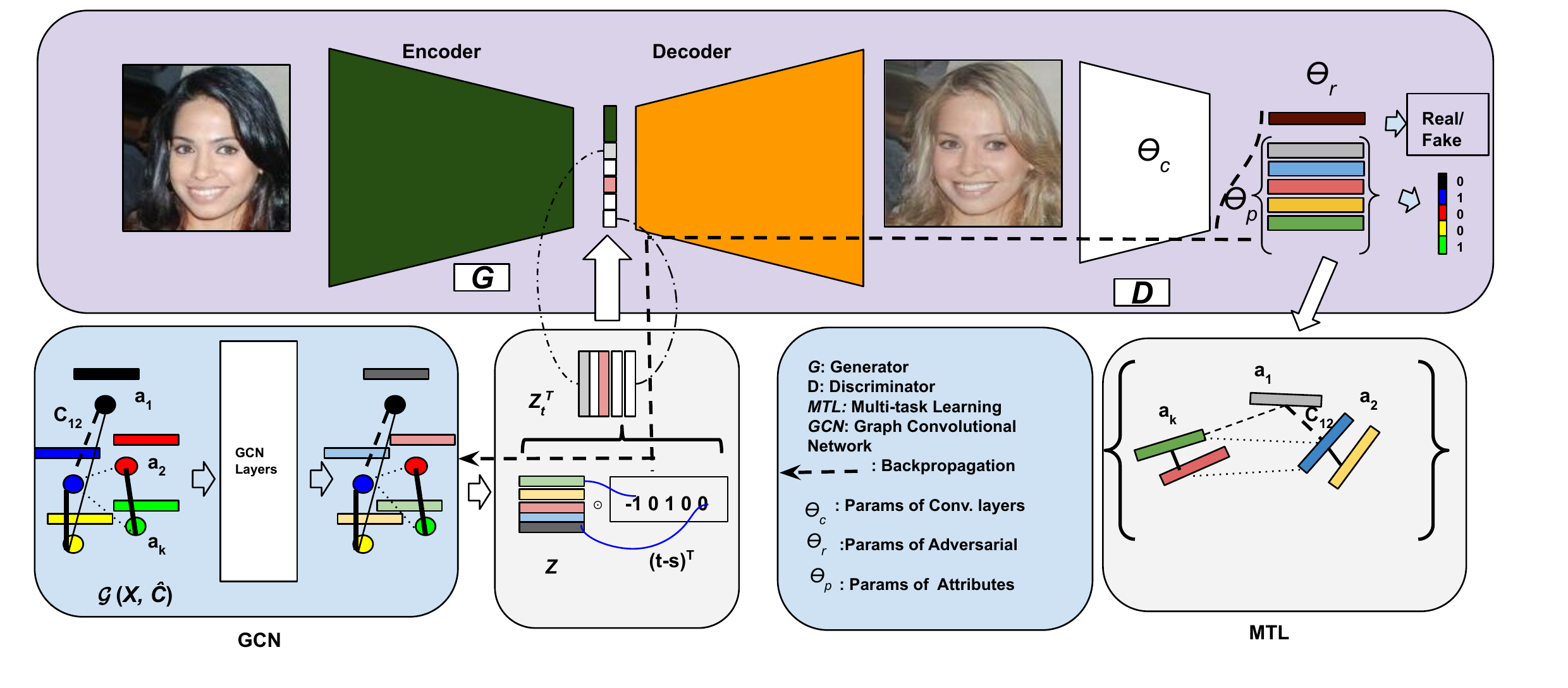}
    \caption{
    Schematic diagram showing the pipeline of the proposed method.  
    Each node of the Graph $\mathcal{G}$ represents an attribute and the edge between them encode their co-occurrence 
    defined on $\widehat{C}$. GCN induces the higher-order representations of the attributes ($Z$) 
    which are further scaled by (t-s) to generate $Z_t$. We concatenate $Z_t$ with the latent representations of input image 
    and feed to the decoder of ($G$). At discriminator, we apply Multi-Task Learning (MTL) to share the weights between 
    the tasks constrained on $C$. During end-to-end learning, we back-propagate the error to the induced representations ($Z$)
    to fine-tune their representations. We maintain the color codes among the attributes (best viewed in color).} 
    \label{fig:main_method}
\end{figure*}

Fig.~\ref{fig:main_method} shows the schematic diagram of the proposed method,
where both the generator $G$ and discriminator $D$ are conditioned.
As mentioned in Sec.~\ref{intro}, existing
cGANs arts such as Stargan~\cite{stargan_cvpr2018}, AttGAN~\cite{he2019attgan} or STGAN~\cite{liu2019stgan} condition the generator, which can be done
either at the encoder or the decoder part of $G$, to synthesise the image with intended attributes. But the problem with their conditions is that
they ignore the intrinsic properties of the attributes and their relationships. They use
single digit (0 or 1) for an attribute, and treat every attributes are equally similar
to each other. In Fig.~\ref{fig:main_method}, the graph on the generator side represents the attributes and their relationships. Each node in the graph represents higher-order semantic representations of attributes,  and the edges between them represent their relationship. 
We propose to induce the attribute representations which encode attribute properties including their relations, based on how they co-occur in the real world scenario. To induce such representations, we propose to apply GCN~\cite{kipf2016semi} with convolutional layers on the generator side. The graph is optimised via end-to-end learning of the
entire system of networks. The discriminator and the attribute classifier guide the graph learning such that the learnt conditional
representations help the generator synthesise more realistic images and preserve target attributes.
Such semantically rich and higher-order conditional representations of the target attributes play an important role in the natural transitioning
to the target attribute. This helps to syntheise images with less artefacts, improved quality and better contrast, as also partially observed in StackGAN~\cite{zhang2017stackgan}.

We also condition the parameters of attributes on the discriminator side using multi-task learning framework, similar to~\cite{cavallanti2010linear}, based on the co-occurrence matrix. Using the learnt representation i.e. the graph to condition the discriminator might also be possible via EM-like alternating optimisation, however due to the complexity and instability, is not considered in this work.
Conditioning both target and its associated attributes on generator and on discriminator enabled GAN to retain the target as well as the associated attributes faithfully~\cite{liu2019attribute}. Unlike~\cite{liu2019attribute} which is hard-coded, limited to a single attribute, 
our method is automatic, and supports arbitrary multiple attributes.
See Section~\ref{sec:multitask} for more details. Before diving into in the details of
the proposed method, we first introduce attributes co-occurrence matrix which is exploited in both the generator and discriminator of the proposed method.

\noindent \textbf{Co-occurrence matrix:} 
To capture the relationship between the attributes based on how frequently they go together, we
constructed a co-occurrence matrix, $C \in \mathbb{R}^{k\times k}$, where $k$ is the total number of attributes. The value at position $(i,j)$
in the matrix gives us an idea about probability of attributes $a_j$ occurring given the attribute $a_i$. Fig.~\ref{fig:coocur_matrix1} shows the 
co-occurrence matrix. We approximate this probability from the training data set as in Eqn.~\ref{eqn:cooccur_mat_prob}. 

\begin{equation}
    P(a_i|a_j) = \frac{\#\mbox{images with $a_i \cap a_j $}}{\#\mbox{images with attribute $a_j$}}
    \label{eqn:cooccur_mat_prob}
\end{equation}

\subsection{Graph Convolution and Generator} 
As stated before, we propose to learn the representations via GCN~\cite{kipf2016semi}, which we simultaneously use to condition the generator. They are in the form of a Graph, $\mathcal{G} = (V, E)$. In our case there are  $k$ different facial attributes, thus total nodes in the graph will be $k$. We represent each node, also called
a vertex $V$ of the graph, by their initial representations of the attributes and the edges between the graph encode their relationship. In our case, this is the co-occurrence probability. Since $P(a_i| a_j) \ne P(a_j|a_i)$, the co-occurrence matrix, $C$ is asymmetric in nature. The graph is constructed from co-occurrence information encoded on $C$ and initial continuous representations of the attributes 
$X \in \mathbb{R}^{k \times d}$. In Fig.~\ref{fig:main_method}, we show a single un-directed edge between the two nodes for clarity. 
The thickness of edges is proportional to the probability of co-occurrence.

The goal of GCN~\cite{kipf2016semi,zhou2018graph} is to learn a function $f(\cdot,\cdot)$ on a graph $\mathcal{G}$, which 
takes initial node representations $X$ and an adjacency matrix ($\widehat{C}$), which we derive from co-occurrence 
matrix  ${C} \in \mathbb{R}^{k \times k}$ as inputs. And, it updates the node features as ${X}^{l+1} \in \mathbb{R}^{k \times d'}$ after 
passing them through every convolutional layer. Every GCN layer can be formulated as
\begin{equation}
    {X}^{l+1} = f({X}^{l}, \widehat{C}) = \sigma({\widehat{C}} {X}^{l} {\theta}^{l}) 
    \label{eqn:graph_cnn_1}
\end{equation}
where $\theta^{l} \in \mathbb{R}^{d \times d'}$ is a transformation matrix learned during training and 
${\widehat{C}} = D^{-1}C \in\mathbb{R}^{k\times{k}}$, $D^{ii}=\sum_{ij}C$ is a diagonal matrix and 
$\sigma$ denotes a non-linear operation, which is LeakyReLU~\cite{maas2013rectifier} for our purpose.
The induced representations of the attributes from the final convolutional layer of GCN, denoted 
as $Z \in \mathbb{R}^{k\times d'}$ are the condition at the generator side as a cue to synthesise new synthetic images. 
Graph convolutions enable combining and averaging information from first-order 
neighbourhood~\cite{kipf2016semi,hamilton2017inductive} to generate higher-order representations of the 
attributes. From Eqn.~\ref{eqn:graph_cnn_1}, we can see that 
the node representations at layer $l+1$ is induced by adding and averaging the representations of a node itself and its 
neighbours from layer $l$. 
The sharing of information from the neighbouring nodes are controlled by the co-occurrence matrix. \\
\noindent \textbf{Generator.} The higher-order representations of the target attributes induced by graph convolution operations
are fed into the generator along with the input image. Recent study~\cite{liu2019stgan} has shown that the difference of target
and source one-hot vector of attributes helps to generate synthetic images with higher rate of target attributes preservation
in comparison to the standard target one-hot vectors~\cite{stargan_cvpr2018,he2019attgan}. We propose to feed the generator
with the graph induced representations of attributes scaled by the difference of target, $t$ and source, $s$ attributes one-hot
vectors 
as: $Z_t = Z\odot(t - s)$,  
where, $Z_t\in \mathbb{R}^{k \times d'}$ is a matrix containing the final representations of the target attributes which 
we feed to the generator as shown in Fig.~\ref{fig:main_method}. 
Given an input image $x$ and the matrix containing continuous representations of target attributes $Z_t$, we learn the parameters,
$\theta_g$ of the generator $G$ to generate a new image $\hat{x}$ in an adversarial manner. The generator usually consists of both
encoder and decoder or only decoder with few convolutional layers. Also conditions are either concatenated with image at encoder
side or concatenated with the image latent representations on the input of decoder side. As we mentioned, our approach is agnostic to the
architectures of GANs. Hence, the induced representations from our approach can be fed into the encoder~\cite{stargan_cvpr2018}
or decoder~\cite{he2019attgan,liu2019stgan} of the generator. In Fig.~\ref{fig:main_method} we present a diagram where 
target attributes conditioning representations are fed from the decoder part of the generator similar to that of
Attgan~\cite{he2019attgan} and STGAN~\cite{liu2019stgan}. We flatten $Z_t$ 
and concatenate it with the latent representations of the input image generated from the encoder and feed it to the decoder. In contrast, in
Stargan~\cite{stargan_cvpr2018} case, each of the columns in $Z_t$ is duplicated separately and overlaid to match the 
dimension $128\times128\times3$ of input RGB image and concatenated with RGB channels.
\[
\hat{x} = G(x,Z_t;\theta_g)
\]

\noindent \textbf{Loss Functions and End-to-end Learning.} The overall loss for the generator is 
$$
\mathcal{L} = \alpha_1\mathcal{L}_{G_{adv}} + \alpha_2\mathcal{L}_{cls} + \alpha_3\mathcal{L}_{rec}
$$
where $\mathcal{L}_{G_{adv}}, \mathcal{L}_{cls}, \mathcal{L}_{rec}$ are the adversarial loss, the classification loss and the reconstruction 
loss respectively and $\alpha_1, \alpha_2, \alpha_3$ represent the hyper-parameters.
We minimise the adversarial loss to make the generated image indistinguishable from the real data. The generator $\theta_g$ and 
discriminator $\theta_d = \{\theta_c, \theta_p, \theta_r\}$ compete to each other in an adversarial manner. Here, $\theta_c$ are the parameters of convolutional layers shared by the discriminator and attribute classifier, $\theta_p$ is of penultimate layers of the classifier, and $\theta_r$ are the parameters of the discriminator. 
In our case, we use WGAN-GP~\cite{arjovsky2017wasserstein,gulrajani2017improved}: 
\[
\begin{split}
\max_{[\theta_c; \theta_r]} \mathcal{L}_{D_{adv}} &= \mathbb{E}_{x}[D(x;[\theta_c, \theta_r])] -           
                                                        \mathbb{E}_{\hat{x}}[D(\hat{x};[\theta_c, \theta_r])]\\  
&+ \lambda \mathbb{E}_{x^{'}=\beta x  + (1-\beta)\hat{x}} [(||\nabla_{x^{'}}D(x^{'}; [\theta_c, \theta_r])||_{2}-1)^{2}] 
\end{split}
\]

\[
\min_{\theta_{g}} \mathcal{L}_{G_{adv}} = \mathbb{E}_{\hat{x}} [1-D(\hat{x}; [\theta_c, \theta_r])]           
\]
where $\hat{x}=G(x,Z_t;\theta_g)$. 

The classification loss here is the standard binary cross-entropy loss in target category label: $
\mathcal{L}_{cls} = \sum_{k}\mathcal{L}^{k}([\theta_c, \theta_p]) 
$, where $\theta_c$ and $\theta_p$ form the attribute classifier. 
The reconstruction loss is computed setting the target attributes equal to that of the source. This results $(t-s)$ into zero vector and 
ultimately,  $Z_t$ turns to the zero matrix. 
\[
\mathcal{L}_{rec} = \| x - G(x, O;\theta_g)\|_{1}           
\]

The above combined loss,~$\mathcal{L}$ trains the generator $G$, the discriminator $D$, and the graph CNN in an end-to-end fashion. The optimal 
attribute representations are learnt to help generate realistic images (by the discriminator loss), and preserve target attributes (by the
classification loss). In the process, multi-attribute relations are also embedded to the representations. 
The networks would consider more natural i.e. realistic when the output image has the presence of associated other attributes as well as the target attribute.

\subsection{Online Multitask Learning for Discriminator}\label{sec:multitask}

While minimising the target attribute classification loss on the discriminator side, we propose to share weights  
between the co-occurring attributes model parameters. We adapted online multitask learning for 
training multiple linear classifiers~\cite{cavallanti2010linear} to achieve this.
The rate of the weights shared between the  model parameters of attributes is constrained by the attributes interaction matrix.
We derive the interaction matrix from the co-occurrence matrix $C$. 

As before, $\theta_c$ are the parameters of convolutional layers and $\theta_p$ is of the penultimate layers of the classifier. We minimize the objective given in Eqn.~\ref{eqn:obj_clas_real} for target attribute classification with respect to discriminator.
The first term in Eqn.~\ref{eqn:obj_clas_real} is the standard 
binary cross entropy loss. The second term is a regularizer which enforces to maintain similar model parameters of frequently co-occurring attributes by sharing the weights. During training, if prediction for any attribute $k$ is wrong, we update not only the parameters of the particular attribute but also the parameters of related attributes. Rate is determined by the co-relation defined on $\widehat{C}$. For more details on regularizer, we suggest to refer the original paper~\cite{cavallanti2010linear}.
\begin{equation}
\mathcal{L}_{cls}^{real} = \sum_{k}\mathcal{L}^{k}([\theta_c, \theta_p]) + \lambda\mathcal{R}(\theta_p, \widehat{C})
\label{eqn:obj_clas_real}
\end{equation}
Note that the multi-task loss is computed on real data. 
Such updates induce similar model parameters of the attributes which are frequently co-occurring as defined in the co-occurrence matrix. 

The multitask attribute classification loss in the above is exploited instead of the conventional single task loss without sharing the parameters. The sharing of parameters between the tasks has advantages over conventional methods: it enforces the discriminator 
to remember the semantic structural relationship between the attributes defined on the co-occurrence matrix. Such kind of 
constrains on the discriminator also helps to minimize the risk of forgetting~\cite{chen2019self} and also retain associated
attributes~\cite{liu2019attribute}. We can also draw analogy 
between our method and Label Smoothing. Our difference from one-sided Label Smoothing~\cite{salimans2016improved} is randomly 
softening the labels while our approach is
constrained with the meaningful co-occurrence matrix and regularises the parameters of the attributes by sharing the weights. 
We train the whole system i.e. $G$, $D$, and the graph in end-to-end as in the previous section, by replacing the binary classification loss with the multitask loss. Conditioning at the discriminator side also helps improve the generator.

\section{Experiments}
\label{experiments}
\noindent \textbf{Data Sets and Evaluation Metrics:} To evaluate the proposed method we carried out our major experiments on 
\textbf{CelebA} which has around $200K$ images annotated with $40$ different attributes. 
In our experiments, we took $13$ attributes similar to that of~\cite{liu2019stgan} on face attribute editing. 
Similarly, \textbf{LFWA} is another benchmark. 
This data set contains around $13,233$ images and each image is annotated with the same 40 different attributes as CelebA. We 
took $12K$ images to train the model and report the performance on the remaining examples. Finally, we use \textbf{RaFD} data set annotated
with the expressions to do attributes transfer from \textbf{CelebA}. This data set consists of $4,824$ images annotated with 
$8$ different facial expressions.

For quantitative evaluations, we employed \textbf{Target Attributes Recognition Rate (TARR)}, \textbf{PSNR}, \textbf{SSIM} 
which are commonly used quantitative metrics for conditional GANs~\cite{stargan_cvpr2018,liu2019stgan,he2019attgan,shmelkov2018good}.
For cGANs, it is not sufficient just to have synthetic realistic images, these being recognisable as the target class is also highly
important~\cite{shmelkov2018good}.  Thus we choose to compute TARR similar to that of existing
works~\cite{stargan_cvpr2018,liu2019stgan,he2019attgan}. TARR measures the generation of conditioned attributes on 
synthetic data by a model trained on real data. 
We took a publicly available pre-trained attribute prediction network~\cite{liu2019stgan} with a mean accuracy of $94.2\%$ on 
$13$ different attributes on test set of CelebA. Similarly, we employ PSNR (Peak Signal to Noise Ratio) and 
SSIM (Structural Similarity) to assess the quality of the synthetic examples. \\
\noindent \textbf{Compared Methods:}
\label{Sec: compared methods} 
To validate our idea, we compare the performance of our GCN induced representations (\emph{gcn-reprs}) with wide ranges of 
both categorical and continuous types of target attributes encoding methods. \\
$\bullet$ \textit{One-hot vector:} As mentioned from the beginning, this is the most commonly and widely used conditioning
technique for cGANs~\cite{stargan_cvpr2018,he2019attgan,lample2017fader,liu2019stgan}. Here, the presence and absence of a
\emph{target attribute} ($t$) is encoded by i.e. $1$ and $0$ respectively.\\
$\bullet$ \textit{Latent Representations (latent-reprs):}~\cite{ding2018exprgan}  
proposed to represent presence/absence of a target expression by a positive/negative $d$-dimensional normal random vector
for expression manipulation. \\
$\bullet$ \textit{Word2Vec:} Words embedding~\cite{mikolov2013distributed} to encode target domain information are successfully applied
to synthesise image from text ~\cite{reed2016generative}. 
We represented target attributes by the embedding of the attributes labels. \\
$\bullet$ \textit{Co-occurrence:} We use co-occurrence vectors as representations of target label attributes to obtain
an approximate performance comparison to~\cite{liu2019attribute}. As,~\cite{liu2019attribute} rules were hard coded 
which is not feasible in our arbitrary attributes manipulation case.\\
$\bullet$ \textit{Attrbs-weights:} We use attribute model parameters obtained from~\cite{liu2019stgan} to represent the target attributes. \\
As~\cite{liu2019stgan} demonstrated the effectiveness of conditioning the difference of target and source attributes one-hot 
vector (t-s) compared to target attributes one-hot vector (t) alone, we mostly perform our experiments on the difference set up. We call conditioning difference of target and source as Difference~\textbf{(Diff)} mode and conditioning target 
only as Standard (\textbf{Std}) mode. We employed several \textbf{\textit{types}} of attribute encoding on both the
\textbf{\textit{modes}} and report their performances on multiple state-of-the-art GAN 
architectures (\textbf{\textit{GAN Archs}}): Stargan~\cite{stargan_cvpr2018}, Attgan~\cite{he2019attgan}, STGAN~\cite{liu2019stgan},
Stargan-JNT~\cite{stargan_cvpr2018}. In addition to these conditioning on the generator side, we also proposed to apply 
Multi-task Learning (\textbf{MTL}) on the discriminator side. \\
\noindent \textbf{Implementation Details:}
We initialise the nodes of the graph with model parameters of attributes (weight vectors) obtained from pre-trained attribute
classifiers~\cite{liu2019stgan}. The dimensions of input and output nodes of the graph are $1024$ and 
$128$ respectively. GCN has 2 convolution layers. For all the data sets in our experiment, we pre-process and crop 
image to the size of $128\times128\times3$.

 \begin{minipage}{\textwidth}
  \begin{minipage}[b]{0.49\textwidth}
    \includegraphics[clip=1.5cm 6cm 1.0cm 1.5cm,width=0.80\linewidth, left]{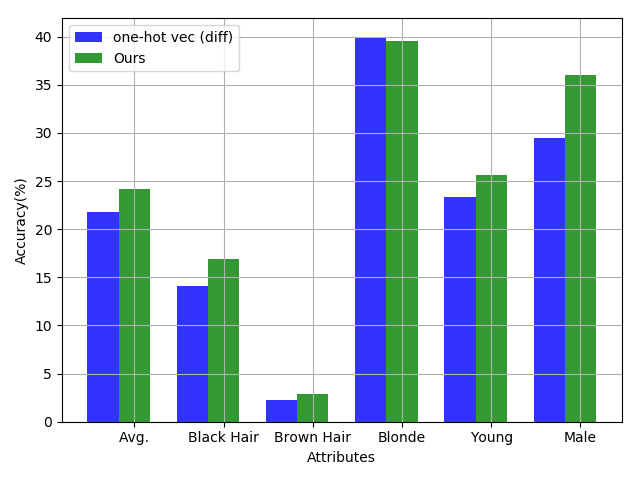}
    \captionof{figure}{Perf. comparison on LFWA}
    \label{fig:tarr_lfwa}
  \end{minipage}
  \hfill
  \begin{minipage}[b]{0.50\textwidth}
    \resizebox{0.9\textwidth}{!}{
    \begin{tabular}{c | c c|c c }
    & \multicolumn{2}{c|}{\textbf{Condition Mode}}  & \\
    \cline{2-3}  
    \rotatebox{0}{\textbf{Condition Type}} & Std   \tab & Diff   &
    \rotatebox{0}{\textbf{Average}}  \\
    \hline
     \textit{one-hot vec} & \checkmark & &	78.6 & \\ 
     \hline
    \textit{one-hot vec} &  & \checkmark  &	80.2 & \\
    \hline
    \textit{co-occurrence} & & \checkmark & 78.6 & \\
    \hline 
    \textit{word2vec} & & \checkmark & 81.3 &  \\
    \hline 
    \textit{attrbs-weights} &  & \checkmark  & 81.9 & \\
     \hline 
     \hline 
    \textit{gcn-reprs}  & & \checkmark & \textbf{84.0} & \\
     \hline 
    \end{tabular} 
    }
     \captionof{table}{Avg. TARR due to various types of attribute representations}
    \label{tab:ablation_attrs_reprs}
    \end{minipage}
  \end{minipage}

\noindent \textbf{Quantitative Evaluations.}\\ 
\noindent \textbf{Ablation Studies:} We train Stargan~\cite{stargan_cvpr2018} 
on both Std and Diff mode with various types of attribute encodings. We computed TARR on 
5 different target attributes viz. hair color~\textit{(black, blond, brown)}, gender~\textit{(male/female)}, 
and age~\textit{(young/old)} on CelebA, similar to the original paper.
Tab.~\ref{tab:ablation_attrs_reprs} summarises the performance comparison. 
Among the four compared condition types (\emph{one-hot vec, co-occurrence, word2vec, attrbs-weights}), \emph{attrbs-weights} obtain the 
best performance. Thus, we chose \emph{attrb-weights} to initialise nodes of GCN. Please note, nodes can be initialised with 
any other type of representations. Referring to the same Table, we observe GCN induced 
representations (\emph{gcn-reprs}) out-performing all other compared methods. \\
\noindent \textit{Discussion:} Semantic representations of attributes: \emph{word2vec} and \emph{attrbs-weights}
outperformed \emph{one-hot vec}. \emph{Co-occurrence} also lagged behind the semantic representations as 
it has no visual information and not optimised for arbitrary attribute manipulations.
Please note,~\cite{liu2019attribute} designed similar representation for single target attribute.
As we know, \emph{word2vec} are learned from the large corpus and bears syntactic and semantic 
relationships between the words~\cite{mikolov2013distributed}. These are also being useful for attributes
classification~\cite{bhattarai2019auglabel}. \emph{Attrbs-weights} are equipped with higher-order visual
information including the spatial location of the attributes (See Fig.~4 from~\cite{taherkhani2018deep}).
Finally, \textit{gcn-reprs} benefited from both semantic representations and co-occurrence relationship.
Thus, it is essential that conditions hold semantically rich higher-order target attributes characteristics.  
\\
\noindent \textit{MTL at discriminator:} We compare our idea to apply MTL at discriminator side. It is evident that, 
in MTL, interaction between the tasks is important. 
Thus, we extend the number of target attributes to $13$ for this experiments. First two rows from the second 
block of  Table~\ref{tab:tarr_perf_comparison} shows the performance comparison between the the baseline w/ and w/o MTL. We observe an overall increase of $+1.1\%$ in performance 
over the baseline. On category level, MTL on the discriminator is outperforming in $9$ different attributes out of $13$ attributes. \\
\noindent \textbf{Comparison With Existing Arts:} To further validate our idea, we compare our method with multiple
state-of-the-art GAN architectures to date in three different quantitative measurements viz. TARR, PSNR and SSIM on 
three different benchmarks. 

\begin{table*}[]
    \renewcommand{\arraystretch}{1.4}
    \centering
    \resizebox{1.0\textwidth}{!}{
    \begin{tabular}{c|c | c c | c c c c c c c c c c c c c|c}
    & & \multicolumn{2}{c|}{\textbf{C. Mode}} &  \multicolumn{13}{c|}{\textbf{Attributes}} & \\  
    \cline{3-17}
    \textbf{\rotatebox{75}{GAN Arch.}} & \textbf{\rotatebox{75}{Condition Type}} &  \rotatebox{0}{Std} & \rotatebox{0}{Diff} & \rotatebox{75}{Bald} & \rotatebox{75}{Bangs} & \rotatebox{75}{Black Hair} & \rotatebox{75}{Blonde Hair} & \rotatebox{75}{Brown Hair} & 
    \rotatebox{75}{B. Eyebrows} & \rotatebox{75}{Eyeglasses} & \rotatebox{75}{Mth. Slt. Open} & \rotatebox{75}{Mustache} & \rotatebox{75}{No beard} & \rotatebox{75}{Pale Skin} 
    & \rotatebox{75}{Male} & \rotatebox{75}{Young} & \rotatebox{75}{Average} \\
   \hline
   \textbf{IcGAN ~\cite{perarnau2016icgan}} & \textit{one-hot vec} & \checkmark &  &  19.4	& 74.2  &	40.6  &	34.6  &	19.7  &	14.7   & 82.4  & 78.8 &	5.5  & 22.6 & 41.8 & 89	& 37.6  &	43.2 \\
  \textbf{FaderNet\cite{lample2017fader}} & \textit{one-hot vec} & \checkmark & &1.5 &	5 & 27  &	20.9 &	15.6 &	24.2 &	87.4 &	44  &	10  &	27.2 &	11.1  &	48.3  &	20.3  &	29.8 \\
    \hline
    \textbf{Stargan~\cite{stargan_cvpr2018}} & \textit{one-hot vec} & \checkmark &  & 24.4 &	92.3 &	59.4 &	68.9 &	55.7 &	50.1 &	95.7 &	96.1 &	18.8 &	66.6 &	84	 & 77.1	 & 83.9	 & 67.2  \\ 
    \textbf{+ MTL} & \textit{one-hot vec} & \checkmark & & 22.7  &	95.4  & 63 & 62.3 &	51.9 &	58  & 99.2 & 98.7 &	24 &	52.2 &	90.5 & 83.7 & 86.8 & 68.3  \\
    & \textit{one-hot vec} & & \checkmark & 41.9  &  93.6 & 	74.7 &	75.2 &	67.4 & 65.9 &	99  &	95.3 &	26.8 &	64.3 &	86.2 &	89 &	89.3 &	74.5 \\
   & \textit{latent-reprs} & \checkmark &  & 18.4  &   93.8 &  68.5 &	60.9 & 62.5	&  69.4 &	97.0  &	97.7 &	14.0 &	 34.4 &	 91.3 &  78.5	 &	 76.7 &	66.4 \\
   &  \textit{latent-reprs} &  & \checkmark & 32.5  &   93.2 &  68.9	 &	79.5 & 71.5	&  55.3 &	97.2  &	98.4 &	30.0 &	 58.5 &	 85.1 &  84.0	 &	 75.1 &	71.4 \\
   &  \textit{attrbs-weights} &  & \checkmark & 32.7  & 96.0 & 74.4 & 77.5 &  74.1 & 66.7 & 98.8 & 32.2 & 78.2 & 90.9 & 81.6 & 98.5 & 86.7 & 76.0 \\
    \cdashline{2-18}
    & \textit{gcn-reprs} &  & \checkmark &	28.2 & \textbf{99.4} & \textbf{76.5}	&  77.1	 & 70.9	 & 74.2 & \textbf{99.5} &	99.4 & 37.3	 & 89.6	 & 92	& 93.4	& 94.9 &	79.4 \\
    \textbf{+ MTL} & \textit{gcn-reprs} &  & \checkmark & 34.4	& 98.4  &	73.3 &	78.6  &	70.8  &	 \textbf{85.5}   & 	\textbf{99.5}  &	99.1  &	44.2  & \textbf{90}	& 92.3 & \textbf{95.6} &	91.7 &	81.0 \\
    \textbf{+ MTL + End2End} & \textit{gcn-reprs} &  & \checkmark & \textbf{56.6} &	98.2 &	76	& \textbf{81.1}	& \textbf{80} &	73.4 &	99.4 & \textbf{99.5} & \textbf{50.9} &	89.1 
    & \textbf{94.2} &	92 &	\textbf{95} &	\textbf{83.4} \\
    \hline 
    \textbf{STGAN~\cite{liu2019stgan}} & \textit{one-hot vec} & \checkmark &  & 40.7 & 	92.5 &	69.5 &	69.7 &	59.4 &	65.2 &	99.2 & 	95	& 26.8 & 69.7 &	90.8 &	70 & 52.8 &	69.3  \\
    & \textit{one-hot vec} &  & \checkmark & 59.9	 & \textbf{97.7}  &	\textbf{93}	& 79 & \textbf{89.9} & \textbf{88.3}	& 99.7  &	96.7 &	38.9 &	93.4 &	\textbf{97.0}  & \textbf{98.5} &	86.7 &	86.1 \\
    \cdashline{2-18} 
    \textbf{+ MTL + End2End} & \textit{gcn-reprs} &  & \checkmark & \textbf{82.0}  & 95.5  & 92.6  & \textbf{85.4}  & 82.0  & 86.2 & \textbf{99.9} & \textbf{99.4} & \textbf{55.3} & \textbf{98.4}  & 96.0  & 98.1  & \textbf{86.9} & \textbf{89.1} \\
    \hline 
    \textbf{AttGAN~\cite{he2019attgan}} & \textit{one-hot vec} & \checkmark &  & 22.5 & 93  &	46.3 &	40.4 & 51 &	49.2 & 98.6	 & 97 & 30.3 &	81.3 &	84.4 &	83.3 &	67.9 &	65 \\
    & \textit{one-hot vec} &  & \checkmark & 69.1  &	97.5 &	78.8 &	 84.4 &	76.5 &	73.4 &	99.6 &	95.8 &	34.2 &	85.8 &	96.8 &	95.8 &	92.9 &	83 \\
    \cdashline{2-18} 
    \textbf{+ MTL + End2End} & \textit{gcn-reprs} &  & \checkmark & \textbf{72.8} & \textbf{97.9} & \textbf{93.9} & \textbf{94.4} & \textbf{92.3} & \textbf{86.8} & \textbf{99.8}  &  \textbf{98.6} & \textbf{48.4} & 
    \textbf{97.1} & \textbf{97.1} & \textbf{98.5} & \textbf{96.4} &   \textbf{90.3}  \\
    \hline 
    \end{tabular}
    }
    \caption{Comparison of Target Attributes Recognition Rate (TARR) on CelebA with different existing cGANs architectures with different target attribute label conditioning.}
    \label{tab:tarr_perf_comparison}
\end{table*}
\noindent \textit{Target Attribute Recognition Rate (TARR):}
Table~\ref{tab:tarr_perf_comparison} compares the TARR on CelebA. 
In the Table, the top block shows the performance of
two earlier works: IcGAN~\cite{perarnau2016icgan} and FaderNet~\cite{lample2017fader}. These figures are as reported on~\cite{liu2019stgan}. These methods relied on \emph{one-hot vec} type of the target attributes in Std mode as their conditions. The TARR of these
methods are modest. 
In the Second block, we compare the performance of \emph{attrbs-weights}, \emph{latent-reprs},\emph{gcn-repres} with 
the default conditioning type of Stargan~\cite{stargan_cvpr2018} on both the conditioning modes. Simply switching 
the default type from Std mode to Diff mode improves 
the TARR from $67.2\%$ to $74.5\%$. This is the best performance of Stargan reported by~\cite{liu2019stgan}.
The encoding principle of~\emph{latent-reprs}~\cite{ding2018exprgan} is similar to that of ~\emph{one hot vec},
as positive and negative random vectors are used instead of 1 and 0. Attribute-specific information rich \emph{attrbs-weights}
out-performs these representations.
We observe the performance of these representations similar to that of \textit{one-hot vec}.
We trained our GCN network to induce~\emph{gcn-reprs} and replaced the default conditioning on the 
generator of Stargan~\cite{stargan_cvpr2018} by \emph{gcn-reprs}, the performance improved to $79.4\%(+4.9\%)$. 
Another experiment with the same set-up on the generator and  MTL on the discriminator improves the performance to $81.0\%(+6.5\%)$.  
Training GCN and GAN on multi-stage fashion make the representations sub-optimal.
Thus, we train GCN and GAN and apply MTL simultaneously and attained an average accuracy of $83.4\%$, the highest performance reported on 
Stargan~\cite{stargan_cvpr2018}. 
We also applied our method to two other best performing GAN architecture~\cite{liu2019stgan,he2019attgan} for 
face attribute manipulations.
Last two blocks of the Table~\ref{tab:tarr_perf_comparison} compares the performance on these architectures. 
Similar to that with Stargan~\cite{stargan_cvpr2018}, instead of existing method of conditioning target attributes and applied ours to both the generator and discriminator and train the model from scratch in an end-to-end fashion.
After we applied our method on STGAN~\cite{liu2019stgan}, which is the state-of-the-art method to date, 
we improve the mean average performance from $86.1\%$ to $89.1\%$.
Similarly, on Attgan~\cite{he2019attgan} we outperformed the best reported performance by $+7.3\%$ and attain the 
new state-of-the-art performance on CelebA. This is $+4.2\%$ above the current state-of-the-art~\cite{liu2019stgan}. 
These results show that our method is agnostic to the cGAN architecture.
We also evaluated our method on LFWA. We applied our method to Stargan and compared it to the default conditioning type, \emph{one-hot vec} on Diff mode.  Fig.~\ref{fig:tarr_lfwa} shows the TARR. From this Fig., 
we can see our approach outperforming the baseline. If we carefully check the performance of individual attributes
on both the benchmarks (Tab.~\ref{tab:tarr_perf_comparison}, Fig.~\ref{fig:tarr_lfwa} ), 
our method is substantially outperforming existing arts in attributes such as \emph{bald, moustache, young, male}. These attributes follows law of nature and it is essential to make natural transition to better retain the target label.

\noindent\textit{PSNR/SSIM:}
We compute PSNR and SSIM scores between real and synthetic images  and compare the performance between with the 
counter-parts. In Tab.~\ref{tab:psnr_comparision}, the two columns \texttt{Before} and \texttt{After} show the 
scores of GANs before and after applying our method respectively.
Our approach consistently improves the performance of the counter-parts. 

\begin{table}[]
     \centering
     \resizebox{0.5\textwidth}{!}{
     \begin{tabular}{*{6}{c|} }
      & \multicolumn{2}{c|}{\textbf{PSNR $\uparrow$}} & \multicolumn{2}{c|}{\textbf{SSIM $\uparrow$}}  &  \\
     \cline{2-5} 
      \textbf{GAN Arch.} & \texttt{Before} & \texttt{After} & \texttt{Before}  &  \texttt{After} &  \textbf{Data Set} \\
      \hline 
      \textbf{Stargan-JNT~\cite{stargan_cvpr2018}} & 23.82 &  \textbf{28.0} & 0.867 &  \textbf{0.944}  & RaFD+CelebA \\ 
      \textbf{StarGAN~\cite{stargan_cvpr2018}} &  22.80 & \textbf{27.20} & 0.819 & \textbf{0.897} & CelebA\\
      \textbf{StarGAN~\cite{stargan_cvpr2018}} &  24.65 & \textbf{27.96} & 0.856 & \textbf{0.912} & LFWA\\
      \textbf{AttGAN~\cite{he2019attgan}}  & 24.07 &  \textbf{26.68} &  0.841 & \textbf{0.858} & CelebA\\ 
      \hline 
     \end{tabular}
     }
     \caption{ Comparison of PSNR and SSIM with existing arts}
     \label{tab:psnr_comparision}
\end{table}


\noindent \textbf{Qualitative Evaluations:}
\begin{figure}
    \centering
    \includegraphics[trim= 0cm 6.2cm 0cm 0.5cm, clip, width=0.90\linewidth]{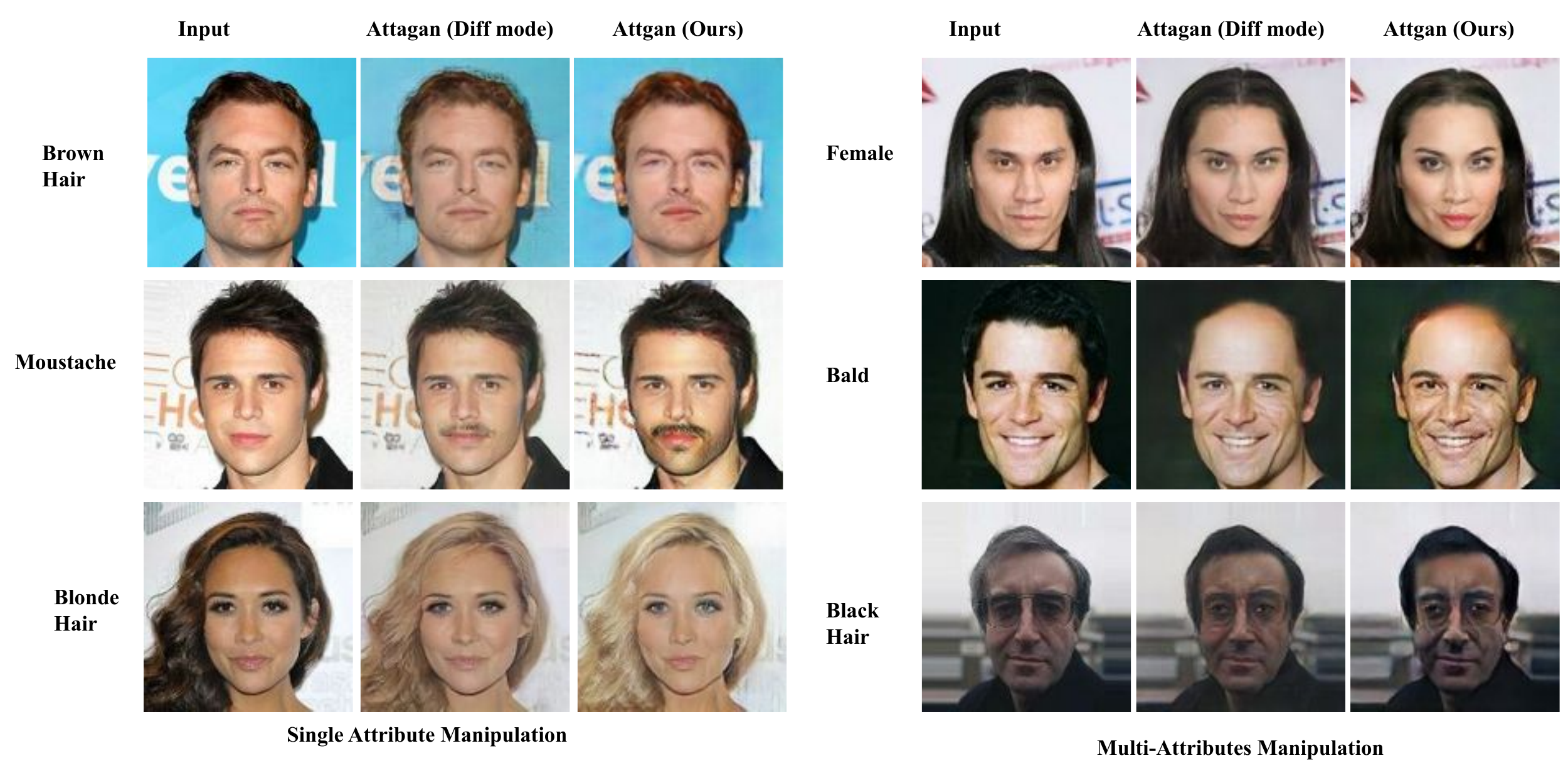}
    \caption{Qualitative comparison of Attgan~\cite{he2019attgan} (default conditioned with \emph{one-hot vector}) trained on diff mode vs Attgan 
    trained with \emph{gcn-reprs (ours)} trained on diff-mode on CelebA (best viewed on color).}
    \label{fig:quali_viz_celeba}
\end{figure}{}
To further validate our idea, we performed extensive qualitative analysis. We compare our method over
the existing arts on two different scenarios i'e in and across data set attributes transfer. \\
\textbf{In Data Set Attribute Transfer:}  In this scenario, we train a model on train set and evaluate the performance 
on the test set of the same data set. Fig.~\ref{fig:quali_viz_celeba} compares the qualitative outcomes of
Attgan~\cite{he2019attgan} conditioned with \emph{one-hot vec} on Diff mode with \emph{gcn-reprs} on the 
same mode. The left block in the figure shows the result of single target attribute manipulation whereas the right block shows 
that of multi-attributes manipulation. From the results, we can clearly see that our method is  able to generate images with 
less artefacts and better  contrast (see the background). In addition to this, our method is  also able to manipulate 
multiple attributes simultaneously, whenever it is meaningful to do so, to give a natural transition from source to target. 
For example, for male-to-female transition, our method is able to put on lipsticks, high cheekbones, arched eyebrows but 
the baseline fails to do so. Similarly, wrinkles on face with few remaining grey hair gives natural transition to bald instead 
just completely removing the hairs from head. As it is highly likely that a person gets bald in his/her old age. Turning grey hair 
to black hair is making the guy comparatively younger as \emph{black hair} is associated with \emph{young} attribute. Due to such
unique strengths of our method, enabled by GCN on the generator and MTL on the discriminator, we observe substantial 
improvements over the baselines especially in the recognition of certain attributes: \emph{Young, Male, Bald, Moustache} where 
a natural transition is essential as these are naturally occurring attributes associated with different factors.

\noindent \textbf{Cross Data Set Attributes Transfer:} 
Stargan-JNT~\cite{stargan_cvpr2018} propose to train a GAN with multiple data sets having disjoint sets of attributes
simultaneously to improve the quality of the cross-data set attribute transfer. We applied our conditioning method  
to train the network on CelebA and RaFD simultaneously. 
\begin{figure}
    \centering
    \includegraphics[trim= 0cm 11cm 0cm 0.0cm, clip, width=0.90\linewidth]{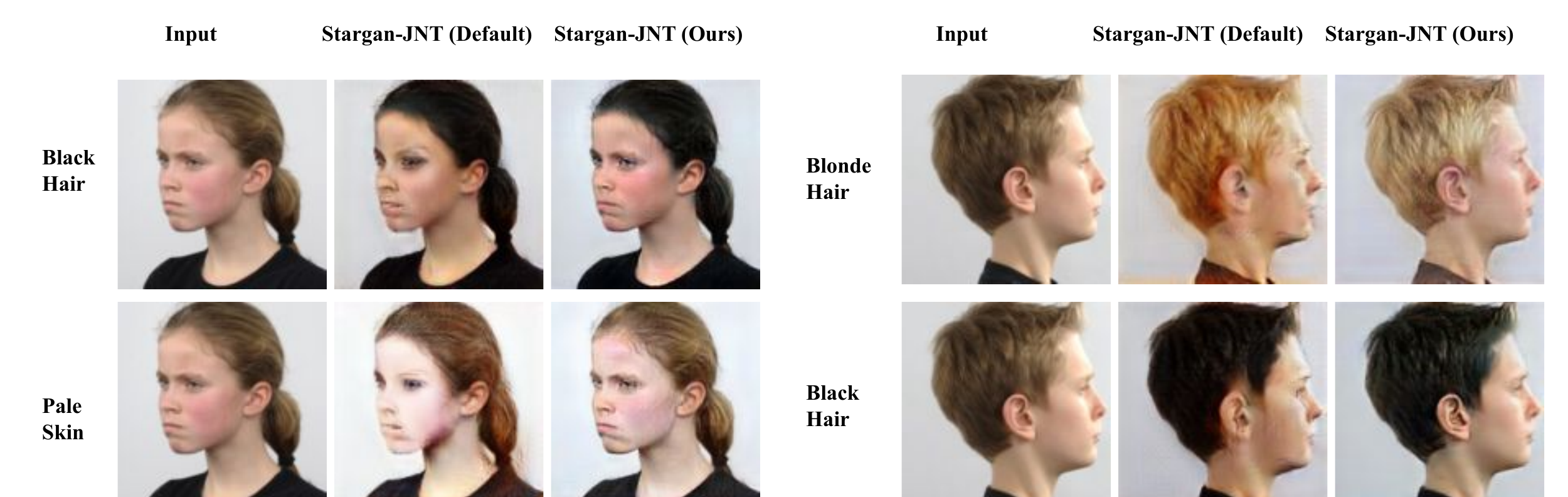}
    \caption{Qualitative comparison of Stargan-JNT with default condition vs Stargan-JNT conditioned with \emph{gcn-reprs (ours)} on RaFD (best viewed on color).}
    \label{fig:qual_viz_rafd}
\end{figure}
And, we compare the performance with their default conditioning method
which is \emph{one-hot vec}. Fig.~\ref{fig:qual_viz_rafd} shows a few test examples from RaFD and their synthetic images when 
target attributes are from CelebA. Please note attribute annotations such as \emph{Black Hair, Blonde Hair, Pale Skin} are 
absent on RaFD train set. From the same Fig., we can clearly see that the synthetic images generated by our method are with less artefacts, 
better contrast and better preservation of the target attributes. 
\section{Conclusions}
\label{conclusions}
We propose a Graph Convolutional Network enabled novel method to induce target attributes embeddings for Conditional 
GANs on the Generator part. Similarly, we proposed a MTL based structural regularisation mechanism on the  
discriminator of the GAN. For both of these, we exploit the co-occurrences of the attributes. Finally, we 
propose a framework to learn 
them in an end-to-end fashion. We applied our method on multiple existing target label conditioned GANs and evaluated on multiple benchmarks for face 
attribute manipulations. From our both extensive quantitative and qualitative evaluations, we observed a substantial 
improvement over the existing arts, attaining new state-of-the-art performance. As a future work, we plan to design 
a framework to dynamically adjust the co-occurrence distribution of the attributes to synthesize naturally 
realistic attributes manipulated images.
\section{Acknowledgements}
Authors would like to thank EPSRC Programme Grant ‘FACER2VM’(EP/N007743/1) for generous support. 
We would also like to thank Prateek Manocha, undergraduate student from IIT Guwahati for some of the baseline 
experiments during his summer internship at Imperial College London.

\bibliographystyle{splncs04}
\bibliography{egbib}

\begin{thebibliography}{10}
\providecommand{\url}[1]{\texttt{#1}}
\providecommand{\urlprefix}{URL }
\providecommand{\doi}[1]{https://doi.org/#1}

\bibitem{arjovsky2017wasserstein}
Arjovsky, M., Chintala, S., Bottou, L.: Wasserstein gan. arXiv preprint
  arXiv:1701.07875  (2017)

\bibitem{bhattarai2019auglabel}
Bhattarai, B., Bodur, R., Kim, T.K.: Auglabel: Exploiting word representations
  to augment labels for face attribute classification. In: ICASSP (2020)

\bibitem{cao2019biphasic}
Cao, J., Huang, H., Li, Y., Liu, J., He, R., Sun, Z.: Biphasic learning of gans
  for high-resolution image-to-image translation. CVPR  (2019)

\bibitem{cavallanti2010linear}
Cavallanti, G., Cesa-Bianchi, N., Gentile, C.: Linear algorithms for online
  multitask classification. JMLR  (2010)

\bibitem{chen2019self}
Chen, T., Zhai, X., Ritter, M., Lucic, M., Houlsby, N.: Self-supervised gans
  via auxiliary rotation loss. In: CVPR (2019)

\bibitem{chen2016infogan}
Chen, X., Duan, Y., Houthooft, R., Schulman, J., Sutskever, I., Abbeel, P.:
  Infogan: Interpretable representation learning by information maximizing
  generative adversarial nets. In: NIPS (2016)

\bibitem{chen2018facelet}
Chen, Y.C., Lin, H., Shu, M., Li, R., Tao, X., Shen, X., Ye, Y., Jia, J.:
  Facelet-bank for fast portrait manipulation. In: CVPR (2018)

\bibitem{chen2019multi}
Chen, Z.M., Wei, X.S., Wang, P., Guo, Y.: Multi-label image recognition with
  graph convolutional networks. In: CVPR (2019)

\bibitem{stargan_cvpr2018}
Choi, Y., Choi, M., Kim, M., Ha, J.W., Kim, S., Choo, J.: {StarGAN: Unified
  generative adversarial networks for multi-domain image-to-image translation}.
  In: CVPR (2018)

\bibitem{ding2018exprgan}
Ding, H., Sricharan, K., Chellappa, R.: Exprgan: Facial expression editing with
  controllable expression intensity. In: AAAI (2018)

\bibitem{gecer2018eccv}
Gecer, B., Bhattarai, B., Kittler, J., Kim, T.K.: Semi-supervised adversarial
  learning to generate photorealistic face images of new identities from 3d
  morphable model. In: ECCV (2018)

\bibitem{grover2016node2vec}
Grover, A., Leskovec, J.: node2vec: Scalable feature learning for networks. In:
  SIGKDD (2016)

\bibitem{gulrajani2017improved}
Gulrajani, I., Ahmed, F., Arjovsky, M., Dumoulin, V., Courville, A.C.: Improved
  training of wasserstein gans. In: NIPS (2017)

\bibitem{hamilton2017inductive}
Hamilton, W., Ying, Z., Leskovec, J.: Inductive representation learning on
  large graphs. In: NIPS (2017)

\bibitem{hamilton2017representation}
Hamilton, W.L., Ying, R., Leskovec, J.: Representation learning on graphs:
  Methods and applications. arXiv preprint arXiv:1709.05584  (2017)

\bibitem{he2019attgan}
He, Z., Zuo, W., Kan, M., Shan, S., Chen, X.: Attgan: Facial attribute editing
  by only changing what you want. IEEE TIP  (2019)

\bibitem{hong2018inferring}
Hong, S., Yang, D., Choi, J., Lee, H.: Inferring semantic layout for
  hierarchical text-to-image synthesis. In: CVPR (2018)

\bibitem{huang2018multimodal}
Huang, X., Liu, M.Y., Belongie, S., Kautz, J.: Multimodal unsupervised
  image-to-image translation. In: ECCV (2018)

\bibitem{isola2017image}
Isola, P., Zhu, J.Y., Zhou, T., Efros, A.A.: Image-to-image translation with
  conditional adversarial networks. In: CVPR (2017)

\bibitem{kaneko2017generative}
Kaneko, T., Hiramatsu, K., Kashino, K.: Generative attribute controller with
  conditional filtered generative adversarial networks. In: CVPR (2017)

\bibitem{kaneko2018generative}
Kaneko, T., Hiramatsu, K., Kashino, K.: Generative adversarial image synthesis
  with decision tree latent controller. In: CVPR (2018)

\bibitem{karras2017progressive}
Karras, T., Aila, T., Laine, S., Lehtinen, J.: Progressive growing of gans for
  improved quality, stability, and variation. ICLR  (2018)

\bibitem{karras2019style}
Karras, T., Laine, S., Aila, T.: A style-based generator architecture for
  generative adversarial networks. In: CVPR (2019)

\bibitem{kipf2016semi}
Kipf, T.N., Welling, M.: Semi-supervised classification with graph
  convolutional networks. In: ICLR (2016)

\bibitem{lample2017fader}
Lample, G., Zeghidour, N., Usunier, N., Bordes, A., Denoyer, L., et~al.: Fader
  networks: Manipulating images by sliding attributes. In: NIPS (2017)

\bibitem{liu2019stgan}
Liu, M., Ding, Y., Xia, M., Liu, X., Ding, E., Zuo, W., Wen, S.: Stgan: A
  unified selective transfer network for arbitrary image attribute editing. In:
  CVPR (2019)

\bibitem{liu2017unsupervised}
Liu, M.Y., Breuel, T., Kautz, J.: Unsupervised image-to-image translation
  networks. In: NeurIPS (2017)

\bibitem{liu2019attribute}
Liu, Y., Li, Q., Sun, Z.: Attribute-aware face aging with wavelet-based
  generative adversarial networks. In: CVPR (2019)

\bibitem{maas2013rectifier}
Maas, A.L., Hannun, A.Y., Ng, A.Y.: Rectifier nonlinearities improve neural
  network acoustic models. In: ICML (2013)

\bibitem{mikolov2013distributed}
Mikolov, T., Sutskever, I., Chen, K., Corrado, G.S., Dean, J.: Distributed
  representations of words and phrases and their compositionality. In: NurIPS
  (2013)

\bibitem{mirza2014conditional}
Mirza, M., Osindero, S.: Conditional generative adversarial nets. arXiv
  preprint arXiv:1411.1784  (2014)

\bibitem{miyato2018spectral}
Miyato, T., Kataoka, T., Koyama, M., Yoshida, Y.: Spectral normalization for
  generative adversarial networks. In: ICLR (2018)

\bibitem{miyato2018cgans}
Miyato, T., Koyama, M.: cgans with projection discriminator. In: ICLR (2018)

\bibitem{nian2019facial}
Nian, F., Chen, X., Yang, S., Lv, G.: Facial attribute recognition with feature
  decoupling and graph convolutional networks. IEEE Access  (2019)

\bibitem{odena2018generator}
Odena, A., Buckman, J., Olsson, C., Brown, T.B., Olah, C., Raffel, C.,
  Goodfellow, I.: Is generator conditioning causally related to gan
  performance? In: ICML (2018)

\bibitem{odena2017conditional}
Odena, A., Olah, C., Shlens, J.: Conditional image synthesis with auxiliary
  classifier gans. In: ICML (2017)

\bibitem{perarnau2016icgan}
Perarnau, G., Van De~Weijer, J., Raducanu, B., {\'A}lvarez, J.M.: Invertible
  conditional gans for image editing. In: NIPSW (2016)

\bibitem{perozzi2014deepwalk}
Perozzi, B., Al-Rfou, R., Skiena, S.: Deepwalk: Online learning of social
  representations. In: SIGKDD (2014)

\bibitem{pumarola2018ganimation}
Pumarola, A., Agudo, A., Martinez, A.M., Sanfeliu, A., Moreno-Noguer, F.:
  Ganimation: Anatomically-aware facial animation from a single image. In: ECCV
  (2018)

\bibitem{reed2016generative}
Reed, S., Akata, Z., Yan, X., Logeswaran, L., Schiele, B., Lee, H.: Generative
  adversarial text to image synthesis (2016)

\bibitem{salimans2016improved}
Salimans, T., Goodfellow, I., Zaremba, W., Cheung, V., Radford, A., Chen, X.:
  Improved techniques for training gans. In: NIPS (2016)

\bibitem{shen2017learning}
Shen, W., Liu, R.: Learning residual images for face attribute manipulation.
  In: CVPR (2017)

\bibitem{shmelkov2018good}
Shmelkov, K., Schmid, C., Alahari, K.: How good is my gan? In: ECCV (2018)

\bibitem{taherkhani2018deep}
Taherkhani, F., Nasrabadi, N.M., Dawson, J.: A deep face identification network
  enhanced by facial attributes prediction. In: CVPRW (2018)

\bibitem{tang2015line}
Tang, J., Qu, M., Wang, M., Zhang, M., Yan, J., Mei, Q.: Line: Large-scale
  information network embedding. In: WWW (2015)

\bibitem{xiao2018elegant}
Xiao, T., Hong, J., Ma, J.: Elegant: Exchanging latent encodings with gan for
  transferring multiple face attributes. In: ECCV (2018)

\bibitem{zakharov2019few}
Zakharov, E., Shysheya, A., Burkov, E., Lempitsky, V.: Few-shot adversarial
  learning of realistic neural talking head models. arXiv preprint
  arXiv:1905.08233  (2019)

\bibitem{zhang2018generative}
Zhang, G., Kan, M., Shan, S., Chen, X.: Generative adversarial network with
  spatial attention for face attribute editing. In: ECCV (2018)

\bibitem{zhang2017stackgan}
Zhang, H., Xu, T., Li, H., Zhang, S., Wang, X., Huang, X., Metaxas, D.N.:
  Stackgan: Text to photo-realistic image synthesis with stacked generative
  adversarial networks. In: CVPR (2017)

\bibitem{zhang2017age}
Zhang, Z., Song, Y., Qi, H.: Age progression/regression by conditional
  adversarial autoencoder. In: CVPR (2017)

\bibitem{zhao2019image}
Zhao, B., Meng, L., Yin, W., Sigal, L.: Image generation from layout. In: CVPR
  (2019)

\bibitem{zhou2018graph}
Zhou, J., Cui, G., Zhang, Z., Yang, C., Liu, Z., Sun, M.: Graph neural
  networks: A review of methods and applications. arXiv:1812.08434  (2018)

\end{thebibliography}
\end{document}